\documentclass{article}

\usepackage{times}
\usepackage{subfigure} 

\usepackage{natbib}

\usepackage{algorithm}
\usepackage{algorithmic}

\usepackage[table]{xcolor}
\usepackage{graphicx}
\usepackage{url}
\usepackage{multirow}
\usepackage{booktabs}
\usepackage{wrapfig}
\usepackage{pbox}
\usepackage{amsmath,amssymb,amsfonts,comment}
\usepackage{subfigure}

\definecolor{coolblack}{rgb}{0.0, 0.18, 0.39}
\definecolor{LinkColor}{rgb}{0,0,0}

\graphicspath{{./figures/}}

\DeclareMathOperator*{\softmax}{softmax}

\usepackage[accepted]{icml2016}

\icmltitlerunning{Ask Me Anything: Dynamic Memory Networks for Natural Language Processing}

\begin{document} 

\twocolumn[
\icmltitle{Ask Me Anything:\\Dynamic Memory Networks for Natural Language Processing}

\icmlauthor{Ankit Kumar, Peter Ondruska, Mohit Iyyer, James Bradbury, Ishaan Gulrajani, Victor Zhong,\\
Romain Paulus, Richard Socher}{}
\icmladdress{firstname@metamind.io, MetaMind, Palo Alto, CA USA}

\icmlkeywords{boring formatting information, machine learning, ICML}

\vskip 0.3in
]

\begin{abstract} 
Most tasks in natural language processing can be cast into question answering (QA) problems over language input. 
We introduce the dynamic memory network (DMN), a neural network architecture which processes input sequences and questions, forms episodic memories, and generates relevant answers. Questions trigger an iterative attention process which allows the model to condition its attention on the inputs and the result of previous iterations. These results are then reasoned over in a hierarchical recurrent sequence model to generate answers. 
The DMN can be trained end-to-end and obtains state-of-the-art results on several types of tasks and datasets: 
question answering (Facebook's bAbI dataset), 
text classification for sentiment analysis (Stanford Sentiment Treebank) and 
sequence modeling for part-of-speech tagging (WSJ-PTB). 
The training for these different tasks relies exclusively on trained word vector representations and input-question-answer triplets.
\end{abstract}

\section{Introduction}
Question answering (QA) is a complex natural language processing task which requires an understanding of the meaning of a text and the ability to reason over relevant facts. Most, if not all, tasks in natural language processing can be cast as a question answering problem: high level tasks like machine translation (\emph{What is the translation into French?}); sequence modeling tasks like named entity recognition \cite{Passos2014} (NER) (\emph{What are the named entity tags in this sentence?}) or part-of-speech tagging (POS) (\emph{What are the part-of-speech tags?}); classification problems like sentiment analysis \cite{Socher2013EMNLP} (\emph{What is the sentiment?}); even multi-sentence joint classification problems like coreference resolution (\emph{Who does "their" refer to?}).

We propose the Dynamic Memory Network (DMN), a neural network based framework for general question answering tasks that is trained using raw input-question-answer triplets. Generally, it can solve sequence tagging tasks, classification problems, sequence-to-sequence tasks and question answering tasks that require transitive reasoning. 

The DMN first computes a representation for all inputs and the question. The question representation then triggers an iterative attention process that searches the inputs and retrieves relevant facts. The DMN memory module then reasons over retrieved facts and provides a vector representation of all relevant information to an answer module which generates the answer. 

Fig.~\ref{fig:example} provides examples of inputs, questions and answers for tasks that are evaluated in this paper and for which a DMN achieves a new level of state-of-the-art performance.

\begin{figure}
\begin{center}
\begin{tabular}{l l }
 I: &  Jane went to the hallway.       \\ 
 I: &  Mary walked to the bathroom.    \\ 
 I: &  Sandra went to the garden.      \\ 
 I: &  Daniel went back to the garden. \\ 
 I: &  Sandra took the milk there.     \\ 
 Q: &  Where is the milk?              \\ 
 A: &  garden                          \\ 
 I: &  It started boring, but then it got interesting.             \\ 
 Q: &  What's the sentiment?           \\ 
 A: &  positive                        \\ 
 Q: &  POS tags?          \\ 
 A: &  PRP VBD JJ , CC RB PRP VBD JJ .
\end{tabular}
\end{center}
\label{fig:example}
\caption{Example inputs and questions, together with answers generated by a dynamic memory network trained on the corresponding task. In sequence modeling tasks, an answer mechanism is triggered at each input word instead of only at the end.}
\end{figure}

\section{Dynamic Memory Networks}
\begin{figure}[t!]
\centering
\includegraphics[width=0.45\textwidth]{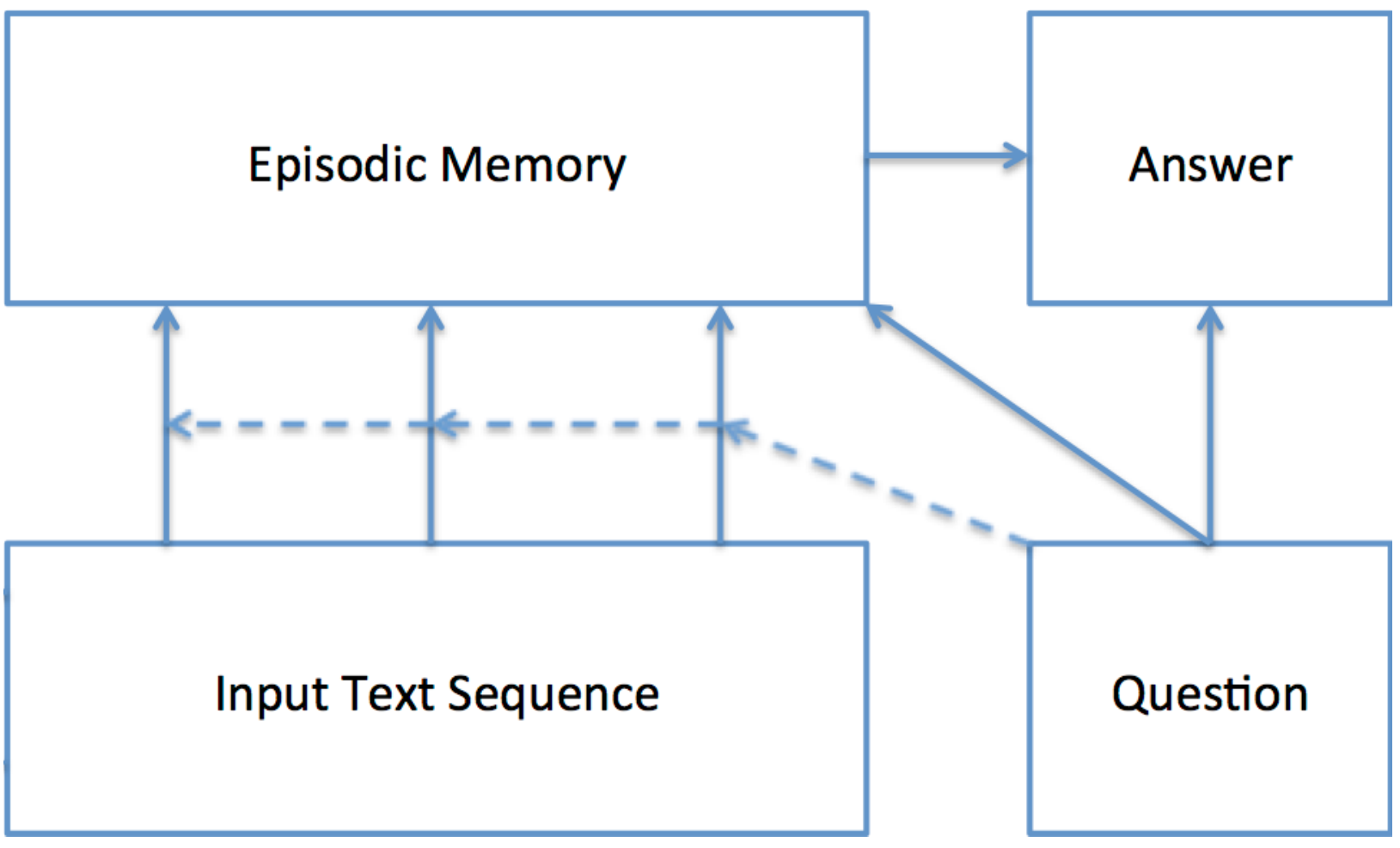}
\label{fig:DMN}
\caption{Overview of DMN modules. Communication between them is indicated by arrows and uses vector representations. Questions trigger gates which allow vectors for certain inputs to be given to the episodic memory module. The final state of the episodic memory is the input to the answer module.}
\vspace{-0.3cm}
\end{figure}

We now give an overview of the modules that make up the DMN. We then examine each module in detail and give intuitions about its formulation. A high-level illustration of the DMN is shown in Fig.~\ref{fig:DMN}.

\textbf{Input Module:}
The input module encodes raw text inputs from the task into distributed vector representations. In this paper, we focus on natural language related problems. In these cases, the input may be a sentence, a long story, a movie review, a news article, or several Wikipedia articles.

\textbf{Question Module:}
Like the input module, the question module encodes the question of the task into a distributed vector representation. For example, in the case of question answering, the question may be a sentence such as \emph{Where did the author first fly?}. The representation is fed into the episodic memory module, and forms the basis, or initial state, upon which the episodic memory module iterates.

\textbf{Episodic Memory Module:}
Given a collection of input representations, the episodic memory module chooses which parts of the inputs to focus on through the attention mechanism. It then produces a "memory" vector representation taking into account the question as well as the previous memory. Each iteration provides the module with newly relevant information about the input. In other words, the module has the ability to retrieve new information, in the form of input representations, which were thought to be irrelevant in previous iterations.

\textbf{Answer Module:} The answer module generates an answer from the final memory vector of the memory module.

\begin{figure*}[t!]
\includegraphics[width=\textwidth]{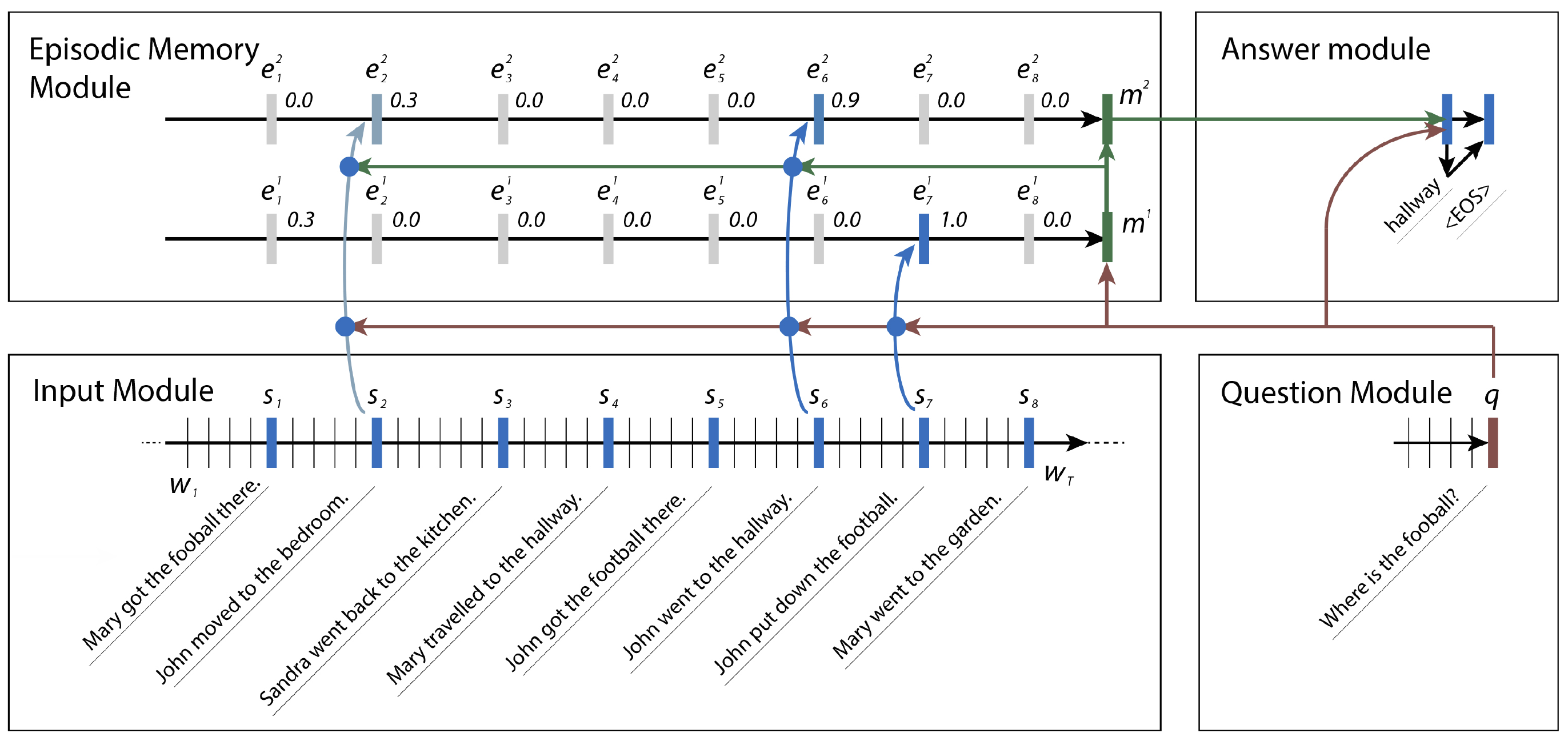}
\caption{Real example of an input list of sentences and the attention gates that are triggered by a specific question from the bAbI tasks \cite{Weston2015ToyTasks}. Gate values $g^i_t$ are shown above the corresponding vectors. The gates change with each search over inputs. We do not draw connections for gates that are close to zero. Note that the second iteration has wrongly placed some weight in sentence 2, which makes some intuitive sense, as sentence 2 is another place John had been.}
\label{fig:ExampleDMN}
\end{figure*}

A detailed visualization of these modules is shown in Fig.\ref{fig:ExampleDMN}.

\subsection{Input Module}\label{section:input}
In natural language processing problems, the input is a sequence of $T_I$ words $w_1,\ldots,w_{T_I}$. One way to encode the input sequence is via a recurrent neural network \cite{Elman1991}. Word embeddings are given as inputs to the recurrent network. At each time step $t$, the network updates its hidden state $h_t = RNN(L[w_t],h_{t-1})$, where $L$ is the embedding matrix and $w_t$ is the word index of the $t$th word of the input sequence.

In cases where the input sequence is a single sentence, the input module outputs the hidden states of the recurrent network. In cases where the input sequence is a list of sentences, we concatenate the sentences into a long list of word tokens, inserting after each sentence an end-of-sentence token. The hidden states at each of the end-of-sentence tokens are then the final representations of the input module. In subsequent sections, we denote the output of the input module as the sequence of $T_C$ fact representations $c$, whereby $c_t$ denotes the $t$th element in the output sequence of the input module. Note that in the case where the input is a single sentence, $T_C = T_I$. That is, the number of output representations is equal to the number of words in the sentence. In the case where the input is a list of sentences, $T_C$ is equal the number of sentences.

\textbf{Choice of recurrent network:}
In our experiments, we use a gated recurrent network (GRU) \cite{Cho2014,Chung2014}. We also explored the more complex LSTM \cite{Hochreiter1997} but it performed similarly and is more computationally expensive. Both work much better than the standard $\tanh$ RNN and we postulate that the main strength comes from having gates that allow the model to suffer less from the vanishing gradient problem \cite{Hochreiter1997}. Assume each time step $t$ has an input $x_t$ and a hidden state $h_t$. The internal mechanics of the GRU is defined as:

\begin{align}
z_t &= \sigma\left(W^{(z)}x_{t} + U^{(z)} h_{t-1}  + b^{(z)} \right)\\
r_t &= \sigma\left(W^{(r)}x_{t} + U^{(r)} h_{t-1} + b^{(r)} \right)\\
\tilde{h}_t &=  \tanh\left(Wx_{t} + r_t \circ U h_{t-1}  + b^{(h)}\right)\\
h_t &=  z_t\circ h_{t-1} + (1-z_t) \circ \tilde{h}_t
\end{align}
where $\circ$ is an element-wise product, $W^{(z)}, W^{(r)}, W \in \mathbb{R}^{n_H \times n_I}$ and $U^{(z)}, U^{(r)}, U \in \mathbb{R}^{n_H \times n_H}$. The dimensions $n$ are hyperparameters. We abbreviate the above computation with $h_t = GRU(x_t,h_{t-1})$.

\subsection{Question Module}

Similar to the input sequence, the question is also most commonly given as a sequence of words in natural language processing problems. As before, we encode the question via a recurrent neural network. Given a question of $T_Q$ words, hidden states for the question encoder at time $t$ is given by $q_t = GRU(L[w^Q_t], q_{t-1})$,  $L$ represents the word embedding matrix as in the previous section and $w^Q_t$ represents the word index of the $t$th word in the question. We share the word embedding matrix across the input module and the question module. Unlike the input module, the question module produces as output the final hidden state of the recurrent network encoder: $q = q_{T_Q}$.

\subsection{Episodic Memory Module}
The episodic memory module iterates over representations outputted by the input module, while updating its internal episodic memory. In its general form, the episodic memory module is comprised of an attention mechanism as well as a recurrent network with which it updates its memory. During each iteration, the attention mechanism attends over the fact representations $c$ while taking into consideration the question representation $q$ and the previous memory $m^{i-1}$ to produce an episode $e^i$. 

The episode is then used, alongside the previous memories $m^{i-1}$, to update the episodic memory $m^{i} = GRU(e^i, m^{i-1})$. The initial state of this GRU is initialized to the question vector itself: $m^0 = q$. For some tasks, it is beneficial for episodic memory module to take multiple passes over the input. After $T_M$ passes, the final memory $m^{T_M}$ is given to the answer module.
 
\textbf{Need for Multiple Episodes:}
The iterative nature of this module allows it to attend to different inputs during each pass. It also allows for a type of transitive inference, since the first pass may uncover the need to retrieve additional facts. For instance, in the example in Fig.~\ref{fig:ExampleDMN}, we are asked \emph{Where is the football?} In the first iteration, the model ought attend to sentence 7 (\emph{John put down the football.}), as the question asks about the football. Only once the model sees that John is relevant can it reason that the second iteration should retrieve where John was. Similarly, a second pass may help for sentiment analysis as we show in the experiments section below.

\textbf{Attention Mechanism:}
In our work, we use a gating function as our attention mechanism. For each pass $i$, the mechanism takes as input a candidate fact $c_t$, a previous memory $m^{i-1}$, and the question $q$ to compute a gate: $g^i_t = G(c_t, m^{i-1}, q)$. 

The scoring function $G$ takes as input the feature set $z(c,m,q)$ and produces a scalar score. We first define a large feature vector that captures a variety of similarities between input, memory and question vectors: $z(c, m, q)=$ 
\begin{equation}
\left[c, m, q, c \circ q, c \circ m, \lvert c - q \rvert, \lvert c - m \rvert, c^T W^{(b)} q, c^T W^{(b)} m \right],
\end{equation}
where $\circ$ is the element-wise product. The function $G$ is a simple two-layer feed forward neural network $G(c,m,q)=$
\begin{equation}
 \sigma\left(W^{(2)} \tanh \left(W^{(1)} z(c, m, q) + b^{(1)}\right) + b^{(2)}\right).
\end{equation}

Some datasets, such as Facebook's bAbI dataset, specify which facts are important for a given question. In those cases, the attention mechanism of the $G$ function can be trained in a supervised fashion with a standard cross-entropy cost function. 

\textbf{Memory Update Mechanism:} To compute the episode for pass $i$, we employ a modified GRU over the sequence of the inputs $c_1,\ldots,c_{T_C}$, weighted by the gates $g^i$. The episode vector that is given to the answer module is the final state of the GRU. The equation to update the hidden states of the GRU at time $t$ and the equation to compute the episode are, respectively:
\begin{eqnarray}
h^i_t &=& g^i_t GRU(c_t, h^i_{t-1}) + (1-g^i_t) h^i_{t-1}\\
e^i &=& h^i_{T_C}\label{eq:ep}
\end{eqnarray}

\textbf{Criteria for Stopping:} The episodic memory module also has a signal to stop iterating over inputs. To achieve this, we append a special end-of-passes representation to the input, and stop the iterative attention process if this representation is chosen by the gate function. For datasets without explicit supervision, we set a maximum number of iterations. The whole module is end-to-end differentiable.

\subsection{Answer Module}\label{section:answer}
The answer module generates an answer given a vector. Depending on the type of task, the answer module is either triggered once at the end of the episodic memory or at each time step.

We employ another GRU whose initial state is initialized to the last memory $a_0 = m^{T_M}$. At each timestep, it takes as input the question $q$, last hidden state $a_{t-1}$, as well as the previously predicted output $y_{t-1}$. 
\begin{eqnarray}
y_t &=& \softmax(W^{(a)} a_t)\\
a_t &=& GRU([y_{t-1},q],a_{t-1}),
\end{eqnarray}
where we concatenate the last generated word and the question vector as the input at each time step.
The output is trained with the cross-entropy error classification of the correct sequence appended with a special end-of-sequence token. 

In the sequence modeling task, we wish to label each word in the original sequence. To this end, the DMN is run in the same way as above over the input words. For word $t$, we replace Eq.~\ref{eq:ep} with $e^i = h^i_t$. Note that the gates for the first pass will be the same for each word, as the question is the same. This allows for speed-up in implementation by computing these gates only once. However, gates for subsequent passes will be different, as the episodes are different.

\subsection{Training}
Training is cast as a supervised classification problem to minimize cross-entropy error of the answer sequence. For datasets with gate supervision, such as bAbI, we add the cross-entropy error of the gates into the overall cost.
Because all modules communicate over vector representations and various types of differentiable and deep neural networks with gates, the entire DMN model can be trained via backpropagation and gradient descent.

\section{Related Work}
Given the many shoulders on which this paper is standing and the many applications to which our model is applied, it is impossible to do related fields justice.
 
\textbf{Deep Learning:}   
There are several deep learning models that have been applied to many different tasks in NLP. For instance, recursive neural networks have been used for parsing \cite{Socher2011}, sentiment analysis \cite{Socher2013EMNLP}, paraphrase detection \cite{Socher2011} and question answering \cite{Iyyer2014} and logical inference \cite{Bowman2014}, among other tasks. However, because they lack the memory and question modules, a single model cannot solve as many varied tasks, nor tasks that require transitive reasoning over multiple sentences.
Another commonly used model is the chain-structured recurrent neural network of the kind we employ above. Recurrent neural networks have been successfully used in language modeling \cite{Mikolov2012}, speech recognition, and sentence generation from images \cite{Karpathy2015}. 
Also relevant is the sequence-to-sequence model used for machine translation by Sutskever et al. \cite{Sutskever2014}. This model uses two extremely large and deep LSTMs to encode a sentence in one language and then decode the sentence in another language. This sequence-to-sequence model is a special case of the DMN without a question and without episodic memory. Instead it maps an input sequence directly to an answer sequence.

\textbf{Attention and Memory:} The second line of work that is very relevant to DMNs is that of attention and memory in deep learning. 
Attention mechanisms are generally useful and can improve image classification \cite{Stollenga2014}, automatic image captioning \cite{Xu2015} and machine translation \cite{Cho2014b,Bahdanau2015}. Neural Turing machines use memory to solve algorithmic problems such as list sorting \cite{Graves2014}. The work of recent months by Weston et al. on memory networks \cite{Weston2015} focuses on adding a memory component for natural language question answering. They have an input (I) and response (R) component and their generalization (G) and output feature map (O) components have some functional overlap with our episodic memory. 
However, the Memory Network cannot be applied to the same variety of NLP tasks since it processes sentences independently and not via a sequence model. It requires bag of $n$-gram vector features as well as a separate feature that captures whether a sentence came before another one. 

Various other neural memory or attention architectures have recently been proposed for algorithmic problems \cite{Joulin2015,Kaiser2015}, caption generation for images \cite{Malinowski2014,Chen2014}, visual question answering \cite{yang2015stacked} or other NLP problems and datasets \cite{Hermann2015}.

In contrast, the DMN employs neural sequence models for input representation, attention, and response mechanisms, thereby naturally capturing position and temporality. As a result, the DMN is directly applicable to a broader range of applications without feature engineering.
We compare directly to Memory Networks on the bAbI dataset \cite{Weston2015ToyTasks}. 

\textbf{NLP Applications:} 
The DMN is a general model which we apply to several NLP problems. We compare to what, to the best of our knowledge, is the current state-of-the-art method for each task. 

There are many different approaches to \emph{question answering}: some build large knowledge bases (KBs) with open information extraction systems \cite{Yates2007}, some use neural networks, dependency trees and KBs \cite{Bordes2012}, others only sentences \cite{Iyyer2014}. A lot of other approaches exist. 
When QA systems do not produce the right answer, it is often unclear if it is because they do not have access to the facts, cannot reason over them or have never seen this type of question or phenomenon. Most QA dataset only have a few hundred questions and answers but require complex reasoning. They can hence not be solved by models that have to learn purely from examples. 
While synthetic datasets \cite{Weston2015ToyTasks} have problems and can often be solved easily with manual feature engineering, they let us disentangle failure modes of models and understand necessary QA capabilities. They are useful for analyzing models that attempt to learn everything and do not rely on external features like coreference, POS, parsing, logical rules, etc. The DMN is such a model.
Another related model by \citet{andreas2016learning} combines neural and logical reasoning for question answering over knowledge bases and visual question answering. 

\emph{Sentiment analysis} is a very useful classification task and recently the Stanford Sentiment Treebank \cite{Socher2013EMNLP} has become a standard benchmark dataset. Kim \cite{Kim2014} reports the previous state-of-the-art result based on a convolutional neural network that uses multiple word vector representations.
The previous best model for \emph{part-of-speech tagging} on the Wall Street Journal section of the Penn Tree Bank \cite{Marcus1993} was Sogaard \cite{Sogaard2011} who used a semisupervised nearest neighbor approach. We also directly compare to paragraph vectors by \cite{Le2014}.

\textbf{Neuroscience:}   
The episodic memory in humans stores specific experiences in their spatial and temporal context. For instance, it might contain the first memory somebody has of flying a hang glider. Eichenbaum and Cohen have argued that episodic memories represent a form of relationship  (i.e., relations between spatial, sensory and temporal information) and that the hippocampus is responsible for general relational learning \cite{Eichenbaum2004}. Interestingly, it also appears that the hippocampus is active during transitive inference \cite{Heckers2004}, and disruption of the hippocampus impairs this ability \cite{Dusek1997}. 

The episodic memory module in the DMN is inspired by these findings. It retrieves specific temporal states that are related to or triggered by a question. Furthermore, we found that the GRU in this module was able to do some transitive inference over the simple facts in the bAbI dataset.
This module also has similarities to the \emph{Temporal Context Model} \cite{Howard2002} and its Bayesian extensions \cite{Socher2009} which were developed to analyze human behavior in word recall experiments.

\section{Experiments}

We include experiments on question answering, part-of-speech tagging, and sentiment analysis. The model is trained independently for each problem, while the architecture remains the same except for the answer module and input fact subsampling (words vs sentences). The answer module, as described in Section \ref{section:answer}, is triggered either once at the end or for each token.

For all datasets we used either the official train, development, test splits or if no development set was defined, we used 10\% of the training set for development. Hyper-parameter tuning and model selection (with early stopping) is done on the development set.
The DMN is trained via backpropagation and Adam \cite{Kingma2014}. 
We employ $L_2$ regularization, and dropout on the word embeddings. Word vectors are pre-trained using GloVe \cite{Pennington2014}. 

\subsection{Question Answering}
The Facebook bAbI dataset is a synthetic dataset for testing a model's ability to retrieve facts and reason over them. Each task tests a different skill that a question answering model ought to have, such as coreference resolution, deduction, and induction. Showing an ability exists here is not sufficient to conclude a model would also exhibit it on real world text data. It is, however, a necessary condition.

\begin{table}[t!]
\begin{center}
\begin{tabular}{l c c } \toprule
    {Task}& {MemNN} & {DMN} \\ \midrule
    1: Single Supporting Fact & 100 & 100  \\
    2: Two Supporting Facts  & {100} & 98.2  \\
    3: Three Supporting Facts  & {100} & 95.2  \\
    4: Two Argument Relations  & 100 & 100  \\ 
    5: Three Argument Relations   & 98 & {99.3} \\
    6: Yes/No Questions  & 100 & 100  \\
    7: Counting  & 85 & {96.9}   \\
    8: Lists/Sets  & 91 & {96.5}  \\ 
    9: Simple Negation  & 100 & 100  \\
    10: Indefinite Knowledge   & 98 & 97.5  \\
    11: Basic Coreference  & 100 & 99.9 \\
    12: Conjunction   & 100 & 100  \\
    13: Compound Coreference   & 100 & 99.8 \\
    14: Time Reasoning   & 99 & 100   \\
    15: Basic Deduction   & 100 & 100 \\
    16: Basic Induction   & 100 & 99.4 \\
    17: Positional Reasoning   & {65} & 59.6 \\
    18: Size Reasoning   & 95 & 95.3 \\
    19: Path Finding   & {36} & 34.5 \\
    20: Agent's Motivations   & 100 & 100 \\
    \midrule
    Mean Accuracy (\%) & 93.3 & \textbf{93.6} \\
    \bottomrule
\end{tabular}
\end{center}
\vspace{-0.3cm}
\caption{Test accuracies on the bAbI dataset. MemNN numbers taken from Weston et al. \cite{Weston2015ToyTasks}. 
The DMN passes (accuracy $>$ 95\%) 18 tasks, whereas the MemNN passes 16.
}
\label{babiresults}
\vspace{-0.3cm}
\end{table}

Training on the bAbI dataset uses the following objective function: $J = \alpha E_{CE}(Gates) + \beta E_{CE}(Answers)$, where $E_{CE}$ is the standard cross-entropy cost and $\alpha$ and $\beta$ are hyperparameters. In practice, we begin training with $\alpha$ set to 1 and $\beta$ set to 0, and then later switch $\beta$ to 1 while keeping $\alpha$ at 1. As described in Section \ref{section:input}, the input module outputs fact representations by taking the encoder hidden states at time steps corresponding to the end-of-sentence tokens. The gate supervision aims to select one sentence per pass; thus, we also experimented with modifying Eq.~\ref{eq:ep} to a simple $\softmax$ instead of a GRU. Here, we compute the final episode vector via:
$e^i = \sum_{t=1}^{T} \softmax(g^{i}_t)c_t$, 
where $\softmax(g^{i}_t) = \frac{\exp(g^{i}_t)}{\sum_{j=1}^T \exp(g^{i}_j)}$, and $g^{i}_t$ here is the value of the gate before the sigmoid. This setting achieves better results, likely because the softmax encourages sparsity and is better suited to picking one sentence at a time.

We list results in Table~\ref{babiresults}. The DMN does worse than the Memory Network, which we refer to from here on as MemNN, on tasks 2 and 3, both tasks with long input sequences.  
We suspect that this is due to the recurrent input sequence model having trouble modeling very long inputs. The MemNN does not suffer from this problem as it views each sentence separately. 
The power of the episodic memory module is evident in tasks 7 and 8, where the DMN significantly outperforms the MemNN. Both tasks require the model to iteratively retrieve facts and store them in a representation that slowly incorporates more of the relevant information of the input sequence. 
Both models do poorly on tasks 17 and 19, though the MemNN does better. We suspect this is due to the MemNN using n-gram vectors and sequence position features.

\subsection{Text Classification: Sentiment Analysis}
The Stanford Sentiment Treebank (SST) \cite{Socher2013EMNLP} is a popular dataset for sentiment classification. It provides phrase-level fine-grained labels, and comes with a train/development/test split. We present results on two formats: fine-grained root prediction, where all full sentences (root nodes) of the test set are to be classified as either very negative, negative, neutral, positive, or very positive, and binary root prediction, where all non-neutral full sentences of the test set are to be classified as either positive or negative.
To train the model, we use all full sentences as well as subsample 50\% of phrase-level labels every epoch. During evaluation, the model is only evaluated on the full sentences (root setup). In binary classification, neutral phrases are removed from the dataset. The DMN achieves state-of-the-art accuracy on the binary classification task, as well as on the fine-grained classification task.

\begin{table}[t!]
\begin{center}
\begin{tabular}{lccc}
\toprule
Task & Binary & Fine-grained \\
\midrule
MV-RNN	& 82.9 & 44.4 \\
RNTN 		& 85.4	& 45.7 \\
DCNN		& 86.8 & 48.5 \\
PVec	    & 87.8 & 48.7 \\
CNN-MC	& 88.1 & 47.4 \\
DRNN		& 86.6 & 49.8 \\
CT-LSTM	& 88.0 & 51.0 \\
\midrule
DMN			& \textbf{88.6} & \textbf{52.1} \\
\bottomrule   
\end{tabular}
\end{center}
\vspace{-0.3cm}
\caption{Test accuracies for sentiment analysis on the Stanford Sentiment Treebank. MV-RNN and RNTN: \citet{Socher2013EMNLP}. DCNN:  \citet{Kalchbrenner2014}. PVec: \citet{Le2014}. CNN-MC: \citet{Kim2014}. DRNN: \citet{Irsoy2015}, 2014. CT-LSTM: \citet{Tai2015}
}
\label{ResultsSentiment}
\vspace{-0.3cm}
\end{table}

In all experiments, the DMN was trained with GRU sequence models. It is easy to replace the GRU sequence model with any of the models listed above, as well as incorporate tree structure in the retrieval process.

\subsection{Sequence Tagging: Part-of-Speech Tagging}
Part-of-speech tagging is traditionally modeled as a sequence tagging problem: every word in a sentence is to be classified into its part-of-speech class (see Fig.~\ref{fig:example}). 
We evaluate on the standard Wall Street Journal dataset \cite{Marcus1993}. We use the standard splits of sections 0-18 for training, 19-21 for development and 22-24 for test sets~\cite{Sogaard2011}.
Since this is a word level tagging task, DMN memories are classified at each time step corresponding to each word. This is described in detail in Section \ref{section:answer}'s discussion of sequence modeling. 

\begin{table}[t!]
\centering
\begin{tabular}{l c c}
 \toprule
 Model & Acc (\%) \\
 \midrule
 SVMTool & 97.15 \\
 Sogaard & 97.27 \\
 Suzuki et al. & 97.40 \\
 Spoustova et al. & 97.44 \\
 SCNN & 97.50 \\
 \midrule
 DMN & \textbf{97.56} \\
 \bottomrule
\end{tabular}
\vspace{-0.3cm}
\caption{Test accuracies on WSJ-PTB}
\label{tab:pos}
\vspace{-0.3cm}
\end{table}

We compare the DMN with the results in \cite{Sogaard2011}. The DMN achieves state-of-the-art accuracy with a single model, reaching a development set accuracy of 97.5. Ensembling the top 4 development models, the DMN gets to 97.58 dev and 97.56 test accuracies, achieving a slightly higher new state-of-the-art (Table~\ref{tab:pos}).

\subsection{Quantitative Analysis of Episodic Memory Module}
The main novelty of the DMN architecture is in its episodic memory module. Hence, we analyze how important the episodic memory module is for NLP tasks and in particular how the number of passes over the input affect accuracy. 

Table \ref{table:episodes} shows the accuracies on a subset of bAbI tasks as well as on the Stanford Sentiment Treebank. We note that for several of the hard reasoning tasks, multiple passes over the inputs are crucial to achieving high performance. For sentiment the differences are smaller. However, two passes outperform a single pass or zero passes. In the latter case, there is no episodic memory at all and outputs are passed directly from the input module to the answer module. We note that, especially complicated examples are more often correctly classified with 2 passes but many examples in sentiment contain only simple sentiment words and no negation or misleading expressions. Hence the need to have a complicated architecture for them is small. The same is true for POS tagging. Here, differences in accuracy are less than 0.1 between different numbers of passes.

Next, we show that the additional correct classifications are hard examples with mixed positive/negative vocabulary.

\subsection{Qualitative Analysis of Episodic Memory Module}
Apart from a quantitative analysis, we also show qualitatively what happens to the attention during multiple passes. We present specific examples from the experiments to illustrate that the iterative nature of the episodic memory module enables the model to focus on relevant parts of the input. For instance, Table \ref{table:attn-babi} shows an example of what the DMN focuses on during each pass of a three-iteration scan on a question from the bAbI dataset.

\begin{table}[t!]
\begin{center}
\begin{tabular}{l cccccccc  c}
\toprule
\pbox{20cm}{Max\\passes} & \pbox{20cm}{task 3\\three-facts} & \pbox{20cm}{task 7\\count} & \pbox{20cm}{task 8\\lists/sets} & \pbox{20cm}{ sentiment\\(fine grain)}\\ 
\midrule
0 pass & 0 & 48.8 & 33.6 & 50.0 \\
1 pass & 0 & 48.8 & 54.0 & 51.5 \\
2 pass & 16.7 & 49.1 & 55.6 & \textbf{52.1} \\
3 pass & 64.7 & 83.4 & 83.4 & 50.1 \\
5 pass & \textbf{95.2} & \textbf{96.9} & \textbf{96.5} & N/A \\
\bottomrule
\end{tabular}
\end{center}
\vspace{-0.3cm}
\caption{Effectiveness of episodic memory module across tasks. Each row shows the final accuracy in term of percentages with a different maximum limit for the number of passes the episodic memory module can take. Note that for the 0-pass DMN, the network essential reduces to the output of the attention module.}
\vspace{-0.3cm}
\label{table:episodes}
\end{table}

\begin{table*}[t!]
\centering
\small
\begin{center}
\textbf{Question:} Where was Mary before the Bedroom? \\
\textbf{Answer:} Cinema.\\
\vspace{0.1in}
\begin{tabular}{l c c c} \toprule
    {Facts} & {Episode 1} & {Episode 2} & {Episode 3} \\ \midrule
    Yesterday Julie traveled to the school. &  &  &  \\
    Yesterday Marie went to the cinema. & & \cellcolor{coolblack!100} &  \\
    This morning Julie traveled to the kitchen. &  &  &  \\
    Bill went back to the cinema yesterday. &  &  &  \\
    Mary went to the bedroom this morning. & \cellcolor{coolblack!100} &  &  \\
    Julie went back to the bedroom this afternoon. &  &  &  \\
    {[done reading]} &  &  & \cellcolor{coolblack!100} \\
    \bottomrule
\end{tabular}
\end{center}
\caption{An example of what the DMN focuses on during each episode on a real query in the bAbI task. Darker colors mean that the attention weight is higher.}
\label{table:attn-babi}
\end{table*}

\begin{figure}[t!]
\vspace{-0.3cm}
  \centering
  \includegraphics[width=0.45\textwidth]{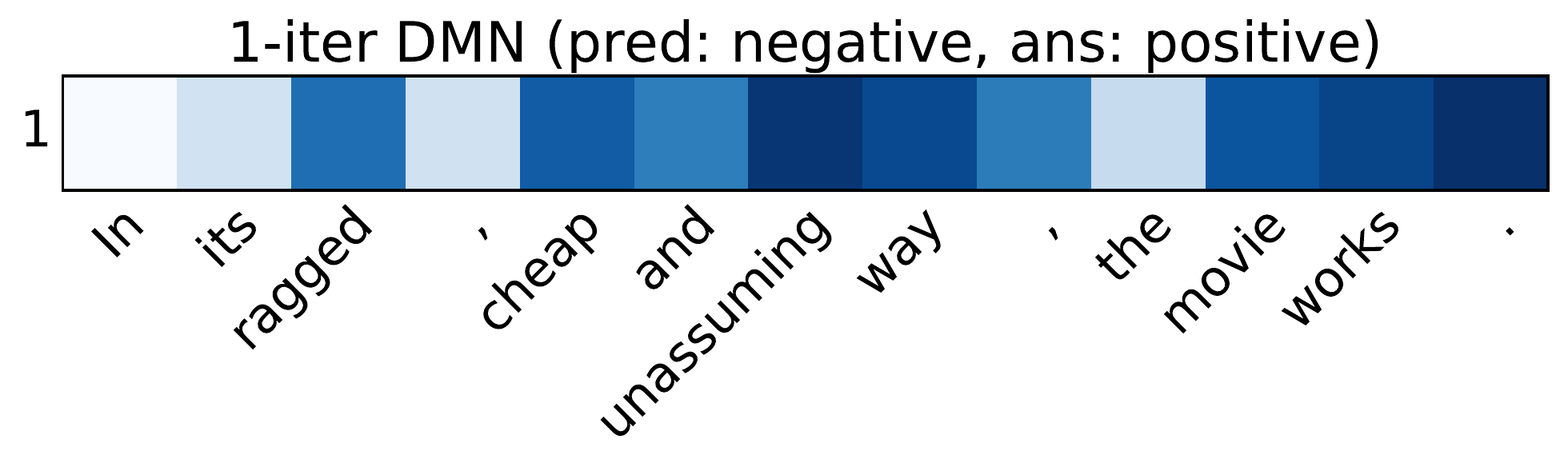}
  \includegraphics[width=0.45\textwidth]{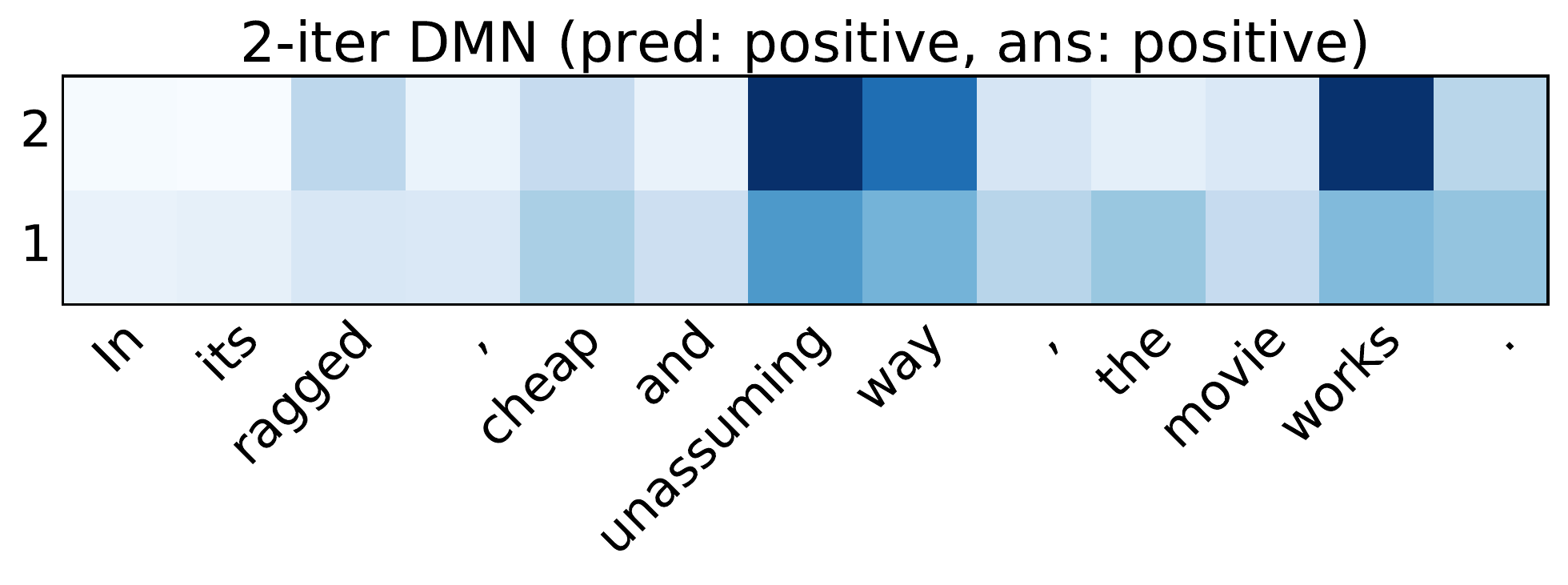}
  \includegraphics[width=0.45\textwidth]{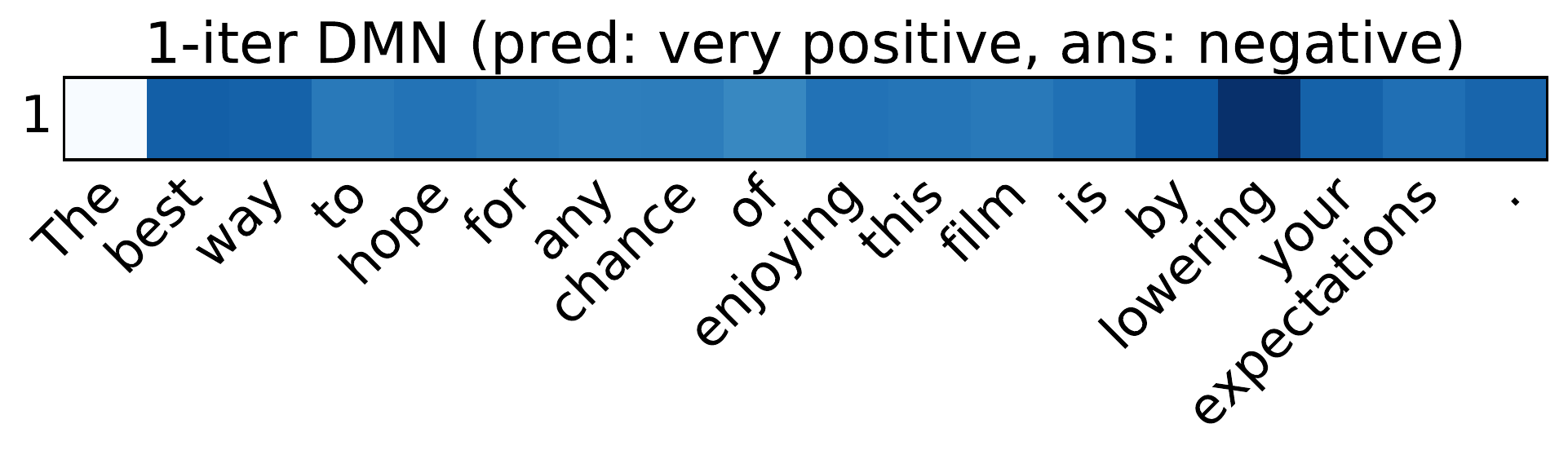}
  \includegraphics[width=0.45\textwidth]{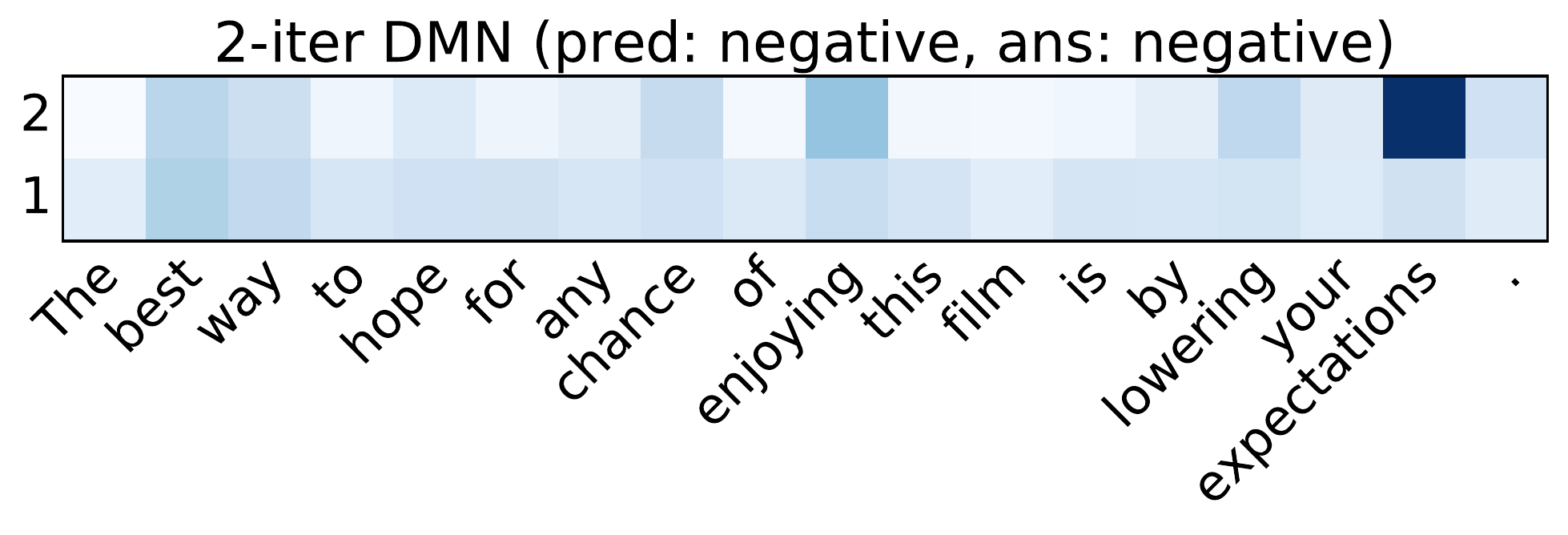}
  \vspace{-0.3cm}
  \caption{Attention weights for sentiment examples that were only labeled correctly by a DMN with two episodes. The y-axis shows the episode number. This sentence demonstrates a case where the ability to iterate allows the DMN to sharply focus on relevant words.}
  \label{sent:focus}
\end{figure}

\begin{figure}[t!]
\vspace{-0.3cm}
  \centering
  \includegraphics[width=0.45\textwidth]{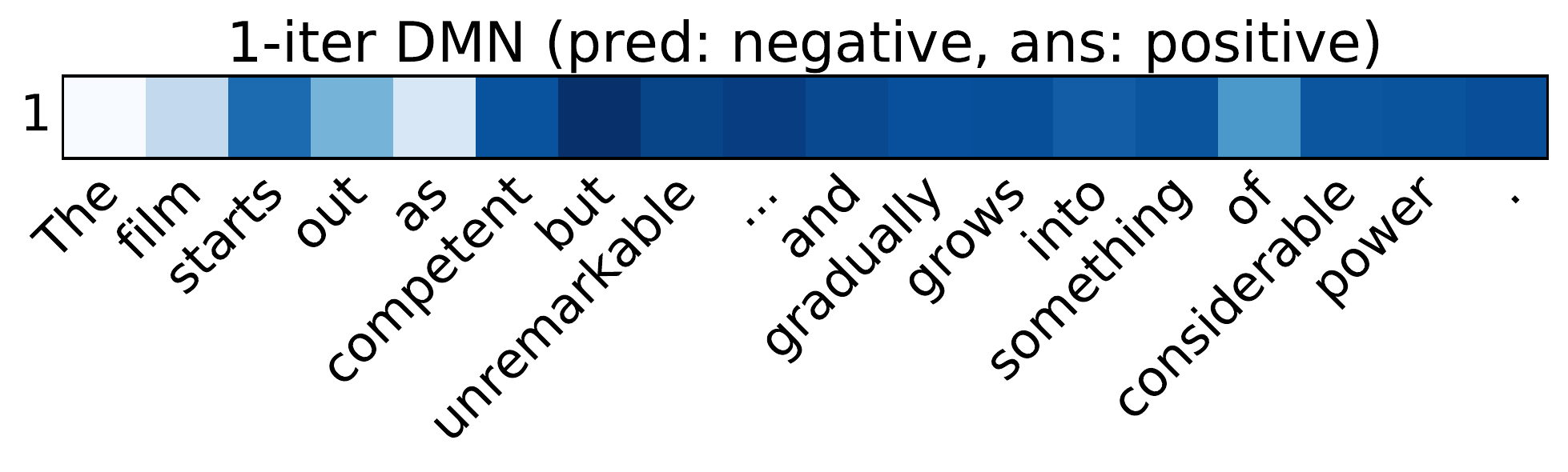}
  \includegraphics[width=0.45\textwidth]{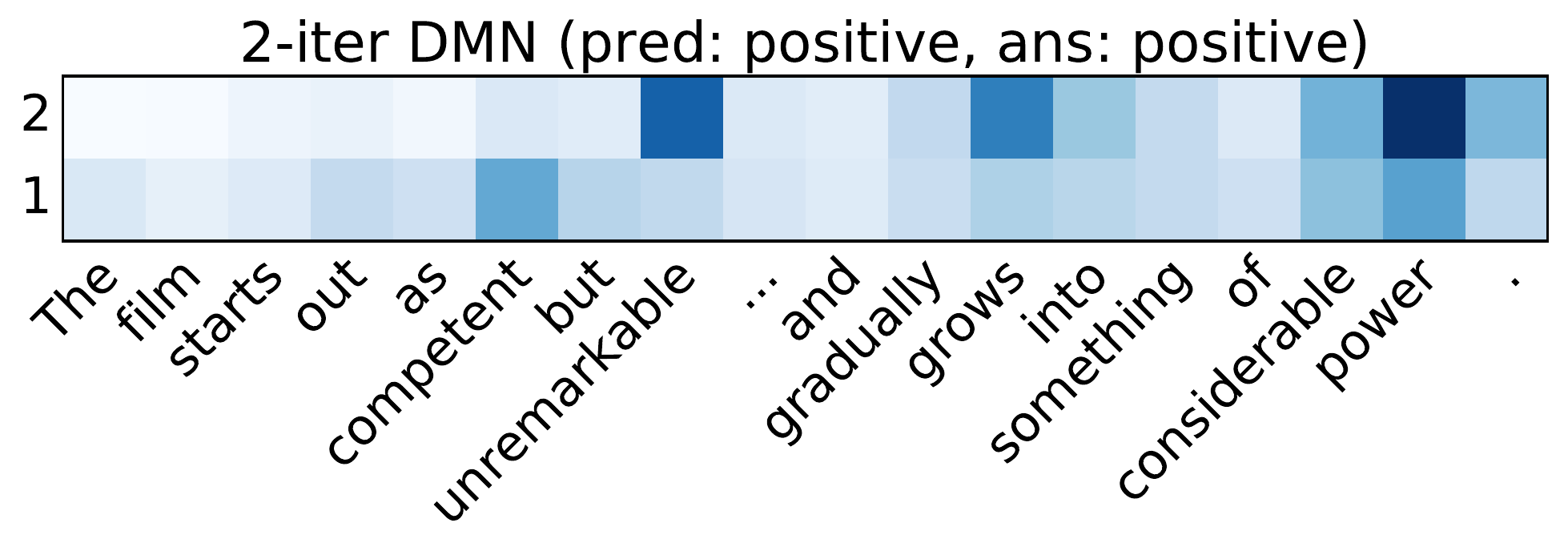}
  \includegraphics[width=0.45\textwidth]{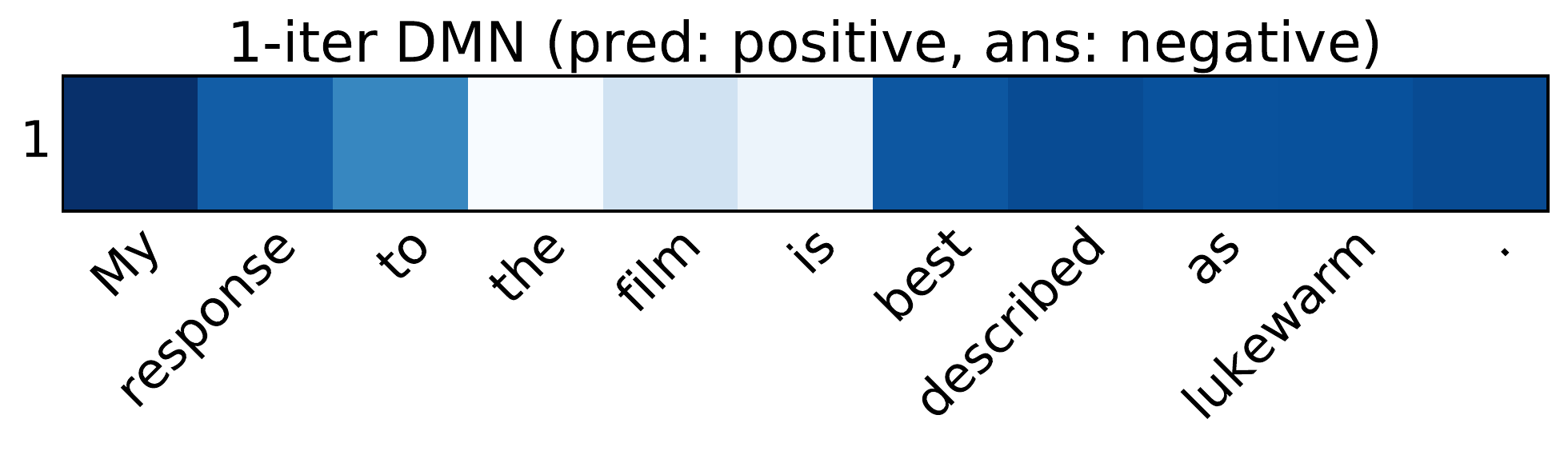}
  \includegraphics[width=0.45\textwidth]{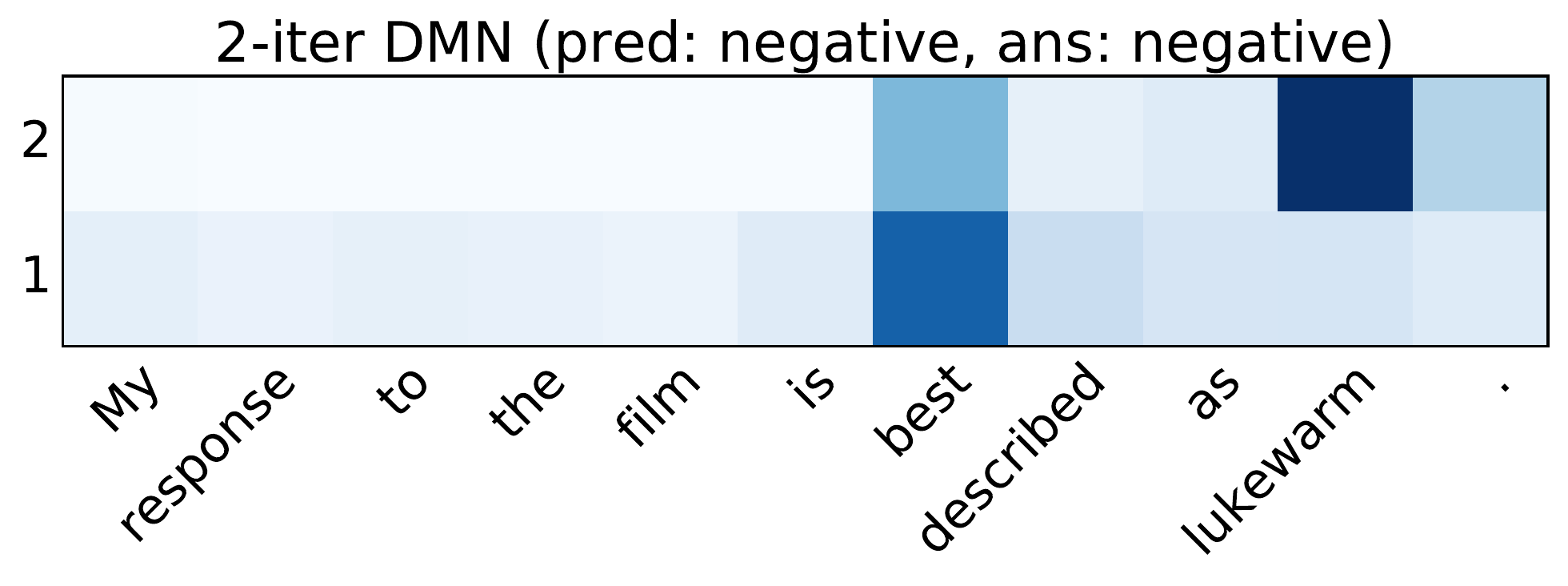}
  \vspace{-0.3cm}
  \caption{These sentence demonstrate cases where initially positive words lost their importance after the entire sentence context became clear either through a contrastive conjunction ("but") or a modified action "best described."}
  \label{sent:neg}
\vspace{-0.4cm}
\end{figure}

We also evaluate the episodic memory module for sentiment analysis. Given that the DMN performs well with both one iteration and two iterations, we study test examples where the one-iteration DMN is incorrect and the two-episode DMN is correct. Looking at the sentences in Fig.~\ref{sent:focus} and \ref{sent:neg}, we make the following observations:
\begin{enumerate}
\item The attention of the two-iteration DMN is generally much more focused compared to that of the one-iteration DMN. We believe this is due to the fact that with fewer iterations over the input, the hidden states of the input module encoder have to capture more of the content of adjacent time steps. Hence, the attention mechanism cannot only focus on a few key time steps. Instead, it needs to pass all necessary information to the answer module from a single pass.
\item During the second iteration of the two-iteration DMN, the attention becomes significantly more focused on relevant key words and less attention is paid to strong sentiment words that lose their sentiment in context. This is exemplified by the sentence in Fig.~\ref{sent:neg} that includes the very positive word "best." In the first iteration, the word "best" dominates the attention scores (darker color means larger score). However, once its context, "is best described", is clear, its relevance is diminished and "lukewarm" becomes more important.
\end{enumerate}

We conclude that the ability of the episodic memory module to perform multiple passes over the data is beneficial. It provides significant benefits on harder bAbI tasks, which require reasoning over several pieces of information or transitive reasoning. Increasing the number of passes also slightly improves the performance on sentiment analysis, though the difference is not as significant. We did not attempt more iterations for sentiment analysis as the model struggles with overfitting with three passes.

\section{Conclusion}
The DMN model is a potentially general architecture for a variety of NLP applications, including classification, question answering and sequence modeling. A single architecture is a first step towards a single joint model for multiple NLP problems. The DMN is trained end-to-end with one, albeit complex, objective function. Future work will explore additional tasks, larger multi-task models and multimodal inputs and questions.


\bibliography{citation}
\bibliographystyle{icml2016}

\end{document}